\documentclass[letterpaper, 10 pt, conference]{ieeeconf}  

 \IEEEoverridecommandlockouts                              

 \usepackage{subfigure}
 \RequirePackage{filecontents}

 \usepackage[utf8]{inputenc} 
 \usepackage[acronym,toc]{glossaries}
  \newacronym{ADF}{ADF}{Assumed Density Filtering}
 \newacronym{B-GPDM}{B-GPDM}{Balanced Gaussian Process Dynamical Models}
 \newacronym{BEV}{BEV}{Bird's Eye View}
 \newacronym{CT}{CT}{Constant Turn}
 \newacronym{CV}{CV}{Constant Velocity}
 \newacronym{CP}{CP}{Constant Position}
 \newacronym{CA}{CA}{Constant Acceleration}
 \newacronym{CNN}{CNN}{Convolutional Neural Network}
 \newacronym{DBN}{DBN}{Dynamic Bayesian Network}
 \newacronym{DN}{DN}{Dense Network}
 \newacronym{FFA}{FFA}{Fuzzy Finite Automata}
 \newacronym{FNR}{FNR}{False Negative Rate}
 \newacronym{EKF}{EKF}{Extended Kalman Filter}
 \newacronym{GAN}{GAN}{Generative Adversarial Network}
 \newacronym{GT}{GT}{Ground Truth}
 \newacronym{GP}{GP}{Gaussian Process}
 \newacronym{GRU}{GRU}{Gated Recurrent Unit}
 \newacronym{GPDM}{GPDM}{Gaussian Process Dynamical Models}
 \newacronym{HMI}{HMI}{Human-Machine Interaction}
 \newacronym{IMM}{IMM}{Interacting Multiple Model}
 \newacronym{ITS}{ITS}{Intelligent Transportation System}
 \newacronym{IMM-EKF}{IMM-EKF}{Interacting Multiple Model Extended Kalman Filter}
 \newacronym{KF}{KF}{Kalman Filter}
 \newacronym{LDS}{LDS}{Linear Dynamical System}
 \newacronym{LRM}{LRM}{Mobile Robotics Laboratory}
 \newacronym{LDCRF}{LDCRF}{Latent Dynamic Conditional Random Fields}
 \newacronym{LSTM}{LSTM}{Long Short-Term Memory}
 \newacronym{MCHOG}{MCHOG}{Motion Contour image based Histogram Of Gradients}
 \newacronym{MLP}{MLP}{Multilayer Perceptron}
 \newacronym{MLE}{MLE}{Maximum Likelihood Estimation}
 \newacronym{PF}{PF}{Particle Filter}
 \newacronym{PHTM}{PHTM}{Probabilistic Hierarchical Trajectory Matching}
 \newacronym{ROC}{ROC}{Receiver Operating Characteristic}
 \newacronym{RMSE}{RMSE}{Root Mean Squared Error}
 \newacronym{RNN}{RNN}{Recurrent Neural Network}
 \newacronym{SVM}{SVM}{Support Vector Machine}
 \newacronym{SLDS}{SLDS}{Switching Linear Dynamical System}
 \newacronym{TPR}{TPR}{True Positive Rate}
 \newacronym{V2I}{V2I}{Vehicle to Infrastructure}
 \newacronym{V2V}{V2V}{Vehicle to Vehicle}
 \newacronym{V2X}{V2X}{Communication between vehicles and anything}
 \newacronym{VR}{VR}{Virtual Reality}
 \newacronym{VRU}{VRU}{Vulnerable Road User}
 \newacronym{WHO}{WHO}{World Health Organization}

 \PassOptionsToPackage{hyphens}{url}\usepackage{hyperref}

 \usepackage{hyperref}
\usepackage{booktabs}
 \usepackage{cite}
 \usepackage{amsmath,amssymb,amsfonts}
 \usepackage{algorithmic}
 \usepackage{graphicx}
 \usepackage{textcomp}

\title{\LARGE \bf
Understanding Pedestrian-Vehicle Interactions with Vehicle Mounted Vision: An LSTM Model and Empirical Analysis
}

\author{Daniela A. Ridel$^{1,2*}$, Nachiket Deo$^{2}$, Denis Wolf$^{1}$, and Mohan Trivedi$^{2}$ 
\thanks{$^{1}$Mobile Robotics Lab, Institute of Mathematics and Computer Sciences, University of Sao Paulo, Brazil {\tt\small danielaridel@usp.br, denis@icmc.usp.br}}%
\thanks{$^{2}$Laboratory for Intelligent and Safe Automobiles,
University of California, San Diego, CA 92092, USA {\tt\small ndeo@ucsd.edu, mtrivedi@ucsd.edu}%
}
\thanks{$^{*}$This work was done when Daniela A. Ridel was a visiting scholar at the Laboratory for Intelligent and Safe Automobiles}%
}

\begin{document}

\maketitle
\thispagestyle{empty}
\pagestyle{empty}

\begin{abstract}
Pedestrians and vehicles often share the road in complex inner city traffic. This leads to interactions between the vehicle and pedestrians, with each affecting the other's motion. In order to create robust methods to reason about pedestrian behavior and to design interfaces of communication between self-driving cars and pedestrians we need to better understand such interactions. In this paper, we present a data-driven approach to implicitly model pedestrians' interactions with vehicles, to better predict pedestrian behavior. We propose a \gls{LSTM} model that takes as input the past trajectories of the pedestrian and ego-vehicle, and pedestrian head orientation, and  predicts the future positions of the pedestrian. Our experiments based on a real-world, inner city dataset captured with vehicle mounted cameras, show that the usage of such cues improve pedestrian prediction when compared to a baseline that purely uses the past trajectory of the pedestrian.

\end{abstract}

\section{Introduction}

The number of fatalities in traffic accidents is currently a major concern in several countries, as stated by the \gls{WHO} \cite{who2}. Most of these accidents are caused by human error. Autonomous vehicles arise as a possible solution to this problem. However, in order to safely and efficiently navigate through inner-city traffic, autonomous vehicles need to predict the intent and motion of surrounding agents. Of particular importance are vulnerable road users such as pedestrians and bicyclists, since their motion is less constrained
compared to vehicles, and the slightest collision could prove fatal.

Pedestrians' intention estimation is the main requirement for safe autonomous navigation. Nowadays pedestrians use eye contact (and possibly gestures) to interact with human drivers as they feel safer to cross the street when the driver sees them. 

\begin{figure}
\includegraphics[width=\columnwidth]{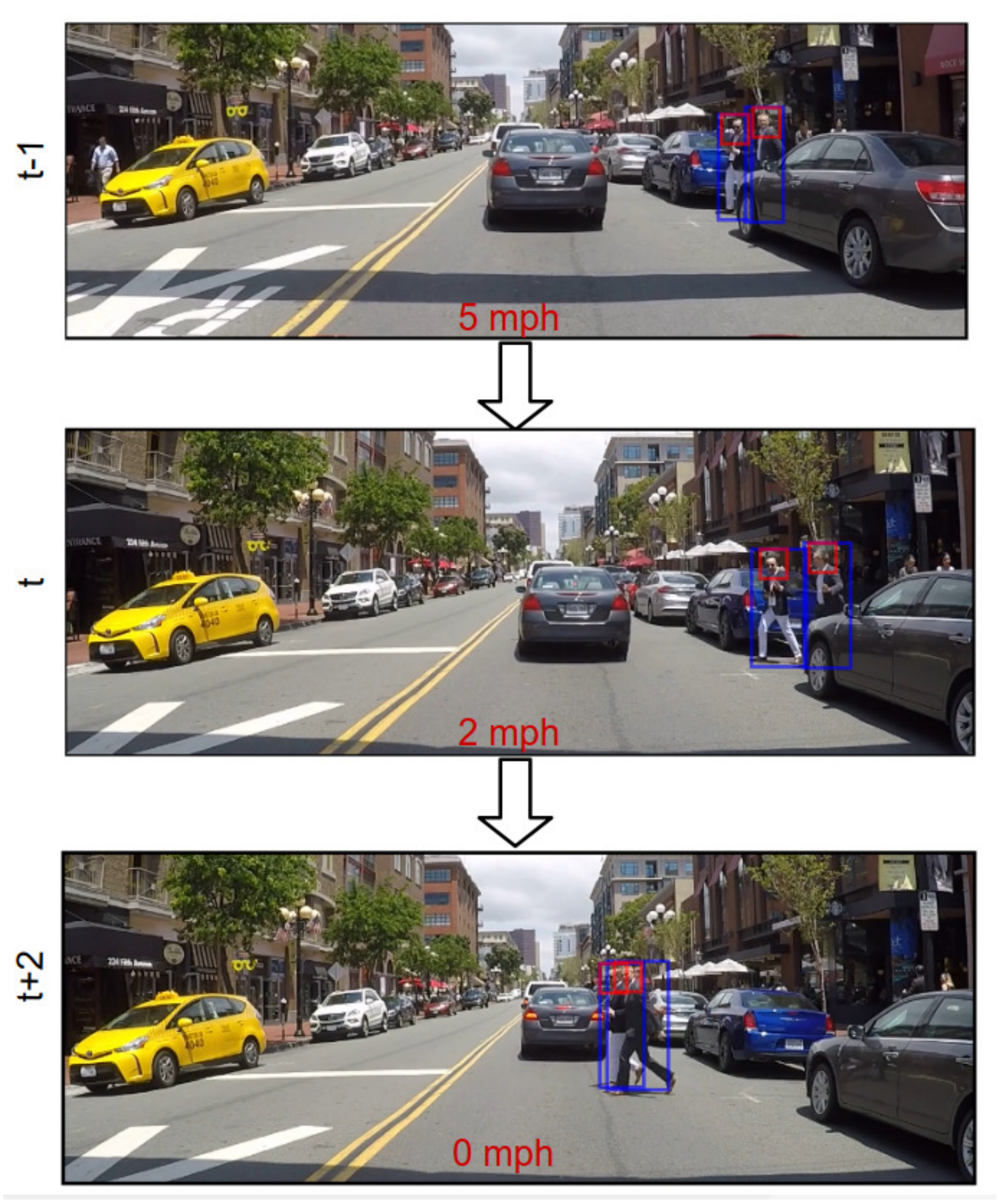}
\caption{Pedestrians and vehicles interact and affect each other's motion. Models for pedestrian intent prediction need to incorporate such interactions. This figure shows one such example. Two pedestrians crossing the street look and gesture at the vehicle. The vehicle then slows down, effectively signaling that it is safe to cross the road. }
\label{fig:intro}
\end{figure}

 The car/pedestrian actions are constantly being affected by each other, i\,.e. a pedestrian may decide to cross a street because he noticed a car speed decrease, and a driver might decide to increase his car speed given he observed the pedestrian is stopping. Pedestrians may also use a gesture to indicate an emergency situation that requires driver's attention. In this context, we are interested in investigating those (and possible others) types of interactions between people and vehicles, from the perception, decision making and interface perspectives.

Pedestrian-vehicle interfaces, especially in the absence of traffic lights, pose a challenging scenario for autonomous vehicles.
Here, both the pedestrian and the vehicle must establish the right of way by predicting each other's intent,
while actively communicating their own, through eye contact, gestures or subtle changes in motion. Figure \ref{fig:intro} shows an example of such interaction between two pedestrians and
the ego-vehicle. We note that the pedestrians make eye-contact and gesture, while moving towards the road from the side-walk.
At the same time, the vehicle slows down, effectively signaling that it is safe for the pedestrians to cross the road. Models for pedestrian behavior prediction can
benefit from learning such interactions between the vehicle and pedestrians.

In this paper, we present a first step towards data-driven approaches for pedestrian behavior prediction, that implicitly learn the interaction between the pedestrian and the ego-vehicle. In particular, our contributions are twofold:
\begin{enumerate}

    \item We propose an LSTM model for pedestrian trajectory prediction that jointly models the motion of the pedestrian based on their past trajectory, the motion of the ego-vehicle based on its past trajectory, and the pedestrian's awareness, based on their head pose.
    \item We evaluate our model using a real world dataset captured using a vehicle mounted camera, involving inner city traffic and unsignalized pedestrian-vehicle interfaces. Our experiments show that jointly modeling the pedestrian's motion and head pose with the ego-vehicle motion leads to lower prediction error, compared to using purely the pedestrian's past trajectory.

\end{enumerate}

\section{Related Work}

\textbf{Pedestrian-vehicle interaction:} 
Interactions among traffic agents have been explored in prior work by means of explicit communication between pedestrians and drivers \cite{Rasouli2018itsc}, pedestrians and cars \cite{Rothenbucher2016,Gupta2018,Chang2017EyesOA}, implicit communication \cite{dey2017}, crosswalks scenarios \cite{schneemann2016,Wang2016}, through the usage of gestures \cite{Gupta2016}, and by analyzing interactions with objects that may interfere with pedestrian awareness of the environment \cite{Rangesh2016}. Recent work presented by Rasouli and Tsotsos \cite{Rasouli2018} summarized pedestrian behavior studies, discussing interactions between pedestrians and autonomous vehicles, and also presented factors that pedestrians take into consideration when crossing streets.
 Pillai et al. \cite{Pillai2017} used \gls{VR} to insert pedestrians in virtual environments. The focus of the study was to estimate the pedestrians' acceptance of autonomous cars. Rothenb\"{u}cher et al. \cite{Rothenbucher2016} used a Wizard of Oz approach to analyze the interactions between pedestrians and autonomous cars, where the driver camouflaged themselves to blend in with the seat, simulating the appearance of autonomous vehicles.

\textbf{Pedestrian trajectory prediction:} Pedestrian trajectory prediction has been extensively addressed in prior research. A detailed review can be found in \cite{Ridel2018}. A majority of proposed models \cite{Alahi2016, kitani2012eccv, socgan, desire, carnet, sophie, vemula2018, varshneya2017, cheng2018, fernando2018 , ballan2016, brendan2018, hasan2018mx} have been evaluated using datasets captured with static cameras, mounted on infrastructure or drones. Since the camera can be treated as a passive observer in these cases, the focus of proposed models has been modeling social (pedestrian-pedestrian) interactions \cite{Alahi2016, socgan, vemula2018, fernando2018, desire, sophie, brendan2018, hasan2018mx}, interactions between scene elements and pedestrians \cite{kitani2012eccv, desire, sophie, carnet, ballan2016, varshneya2017} or more recently interactions in mixed traffic involving pedestrians and vehicles \cite{cheng2018}. Contrary to these approaches, we consider pedestrians observed using vehicle mounted cameras. Thus, the observed trajectories are strongly affected by the ego-vehicle motion. Additionally, the first person perspective allows for a finer-grained analysis of pedestrian activity, including pedestrian gaze or gestures.

More closely related to our work are approaches that use vehicle mounted cameras and use information of pedestrian dynamics coupled with the pedestrian's awareness \cite{Schulz2015, Schulz2015b,Kooij2014,Hashimoto2015b,Hashimoto2015,Keller2014}. However they are usually focused on specific scenarios as signalized crosswalks \cite{Hashimoto2015b} and intersections \cite{Hashimoto2015}, or does not have a high diversity in number of scenarios and pedestrians \cite{Kooij2014, Keller2014,Schulz2015,Schulz2015b}. Our approach is not restricted to a specific scenario and we reason about pedestrian future positions by learning such cues from realistic data.

\section{Motivation}

Autonomous cars must detect pedestrians and discover if they are trying to cross a road, or if they are just standing or moving along the sidewalk. When the car detects that the pedestrian has the intention to cross the street. The car must inform to him that it is possible to do so. Here we can see basically three problems: detecting the pedestrians, predicting their movements and building an efficient interface of communication with them. In this work we focus on the second problem that is the prediction of pedestrian behavior.

In this work we study the influence of ego vehicle dynamics in pedestrian motion, and how both affect themselves when making decisions on crossing streets. As suggested by prior work that pedestrians seek eye contact before crossing a street \cite{semconreport}, we have also incorporated pedestrian head orientation to see in a data-driven approach if such cues interferes with pedestrian future trajectory.

There is an implicit type of communication between cars and pedestrians which has not been extensively studied as pedestrians strong rely on cues given by cars in order to make decisions when walking on streets. Nowadays, when pedestrians want to cross a street, they stare at the vehicle's conductor to be sure that they have been seen. If the answer is positive; they feel confident to cross the street.

At every moment cars and pedestrians are making new decisions regarding all the information being updated due to changes in the environment. Pedestrians  may decide to not cross a street because the street is too large, or due to a car that is approaching at high speed, or because all other pedestrians are waiting. But given only one action all other agents may update their behavior and perform another action, e.g\,. when a pedestrian steps to cross a street and suddenly all other pedestrians follow him, or if a car starts to decrease speed and all pedestrians waiting are encouraged to cross. This basically means that the actions of different agents are somehow correlated and should be used in order to predict each agent future action. One of those interactions happens between cars and pedestrians by means of their motion.

\section{Method}
\begin{figure*}[t]
\centering
\includegraphics[width=\textwidth]{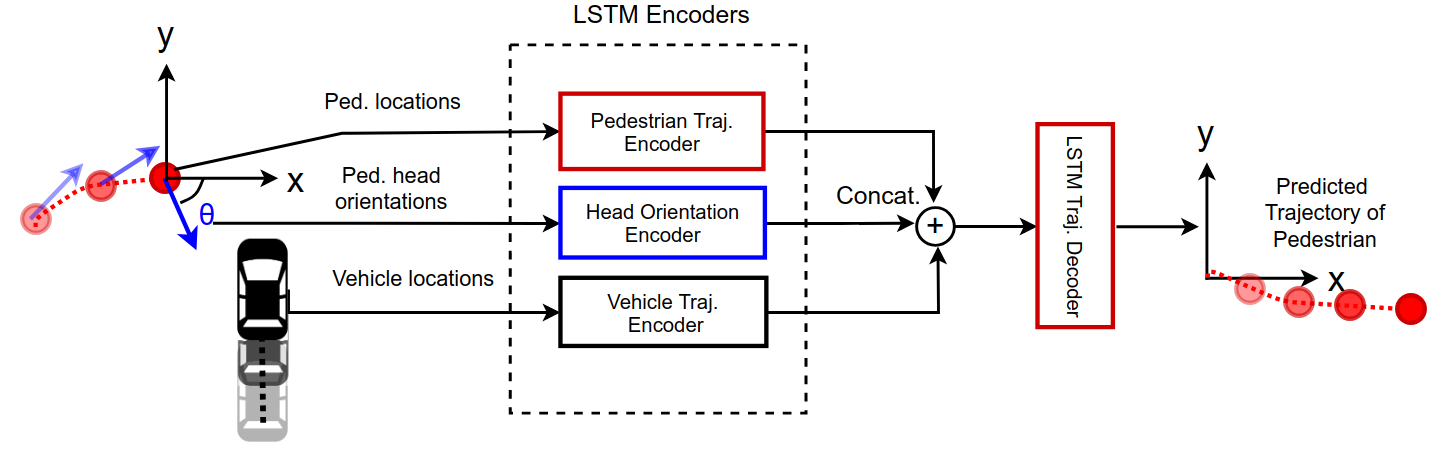}
\caption{\textbf{Overview of proposed model:} We encode the past trajectories of the pedestrian and vehicle, and the head orientations of the pedestrian over 1 second intervals using LSTM encoders. The final state of each LSTM encoder is concatenated and passed to an LSTM decoder. The decoder generates the predicted trajectory of the pedestrian over a prediction horizon of 2 seconds.}
\label{fig:netEncoderDecoderN}
\end{figure*}

We encoded pedestrian and ego-vehicle data in an Encoder/Decoder network based on the work of \cite{Deo2018}. The network comprises an encoder-decoder architecture based on LSTM blocks. As we focused on the interaction between ego-vehicle and pedestrian we did not use the social pooling layer presented on the original paper. The model encodes pedestrian positions, head orientation and ego-vehicle locations in LSTM encoders, concatenating the outputs and using them as input to an LSTM decoder that outputs pedestrian future positions, as depicted in Fig. \ref{fig:netEncoderDecoderN}. The input of the network was one second of information with intervals of $0.2$ seconds, $\Delta t$.
$$
X = [x^{t-4\Delta t},...,x^{t-\Delta t},x^{t}]
$$

Each position value used as input was an x-y coordinate in the pedestrian at $t$ position frame of reference

$$
x^{t} = [x^{t}_{pos},y^{t}_{pos}]
$$

The output of this network is 2 seconds of pedestrian future positions (x and y) also with the same interval $\Delta t$ of time.

$$
Y = [x^{t+\Delta t},x^{t+2\Delta t},...,x^{t+10\Delta t}]
$$

As $X$ inputs we used pedestrian and ego-vehicle positions ($X_{ped}$ and $X_{ego}$). We also used as input the head orientation of the pedestrian.
$$
\Theta_{ped} = [\theta_{ped}^{t-4\Delta t},...,\theta_{ped}^{t-\Delta t},\theta_{ped}^{t}]
$$

The model was implemented using PyTorch \cite{Paszke2017pytorch} and we followed PyTorch recommendations available in: \cite{manualpytorch} to get reproducible results and we maintained the same random initialization of the weights for all the experiments, as the data was not big enough the random initialization of the weights caused small variations on the output. We have also used Adam \cite{kingma2014adam} as optimizer and Mean Squared Error as loss function.

\section{Experiments}

\begin{figure*}
\centering
\subfigure[Pedestrian (red bounding box) seeks eye contact with the driver, driver stops and the pedestrian crosses the street. Exemplification of a scenario where pedestrian and vehicle affected each other's actions.]{\includegraphics[width=0.65\columnwidth]{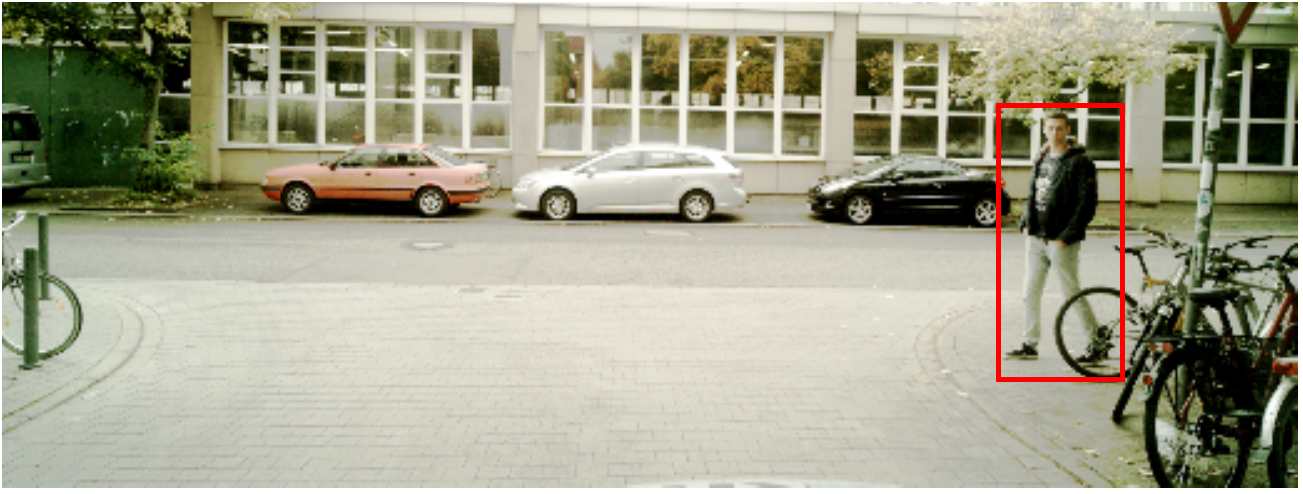}\label{fig:seekapprovalfig}}
\quad
\subfigure[Pedestrian (red bounding box) walks in direction of the ego-lane, driver does not stop, and the pedestrian stops to avoid a collision with the ego-vehicle. Exemplification of a scenario where pedestrian and vehicle affected each other's actions.]{\includegraphics[width=0.65\columnwidth]{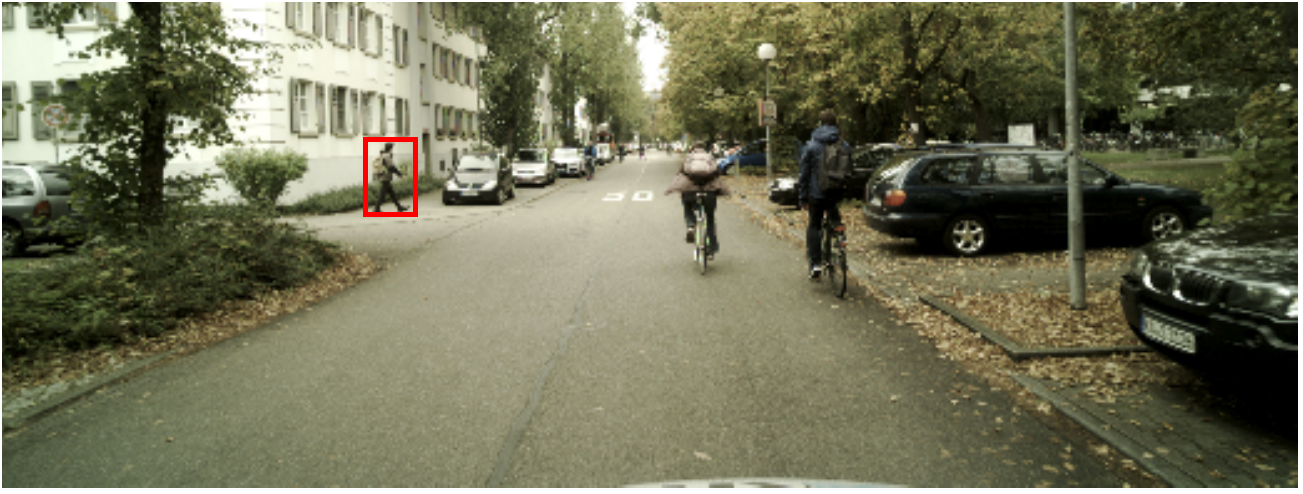}\label{fig:pedstopfig}}
\quad
\subfigure[Pedestrian (red bounding box) crosses the street with near constant speed, and ego-vehicle keeps near constant speed. Exemplification of a scenario where pedestrian and vehicle did not affect each other's actions, probably because of the high distance.]{\includegraphics[width=0.65\columnwidth]{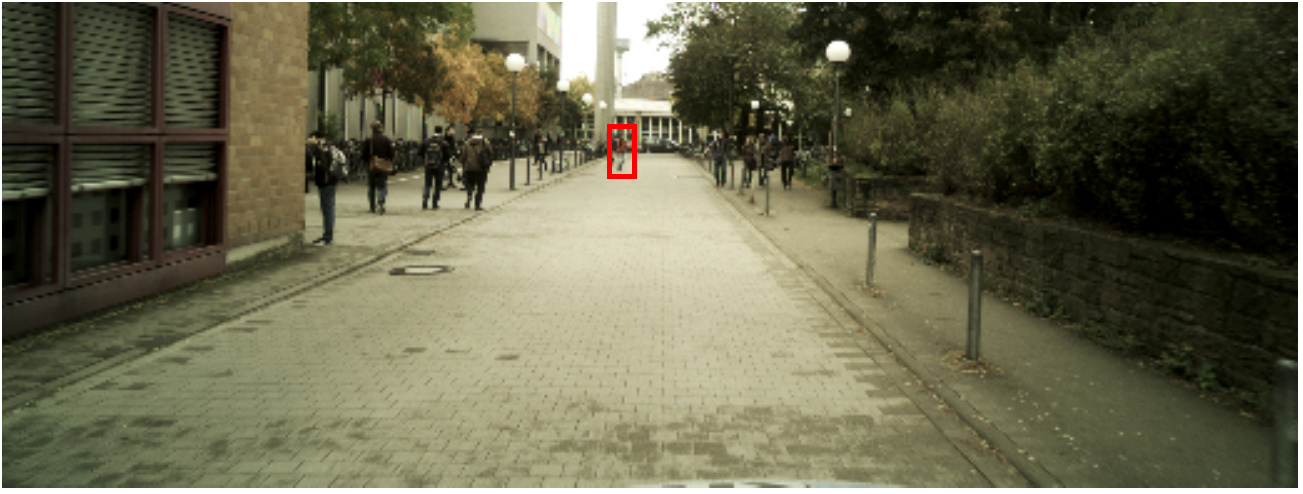}\label{fig:highdistanceped}}
\qquad
\subfigure[Pedestrian keeps eye contact with the driver (third graph), ego-vehicle (green line) decreases its own speed and pedestrian (red line) increases his speed (first graph).]{\includegraphics[width=0.65\columnwidth]{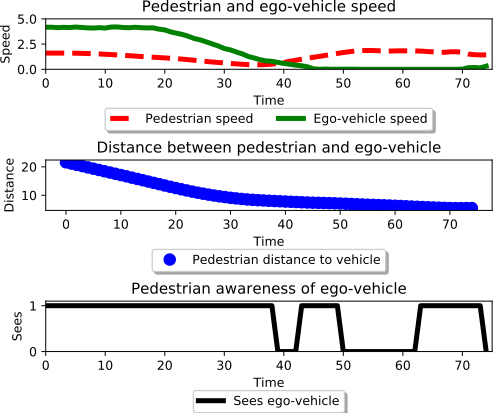}\label{fig:seekapprovalfigb}}
\quad
\subfigure[The ego-vehicle speed (green line) keeps near constant whether pedestrian speed (red line) decreases until reaching zero (first graph).]{\includegraphics[width=0.65\columnwidth]{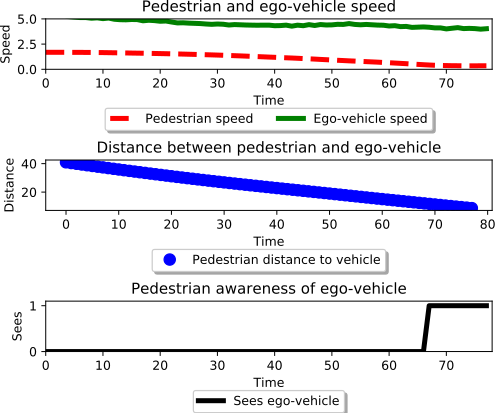}}
\quad
\subfigure[Pedestrian and ego-vehicle are at a considerable distance and both keep constant velocity (first graph). We noticed that in such scenarios their decisions tend to not have much impact in each other's decisions.]{\includegraphics[width=0.65\columnwidth]{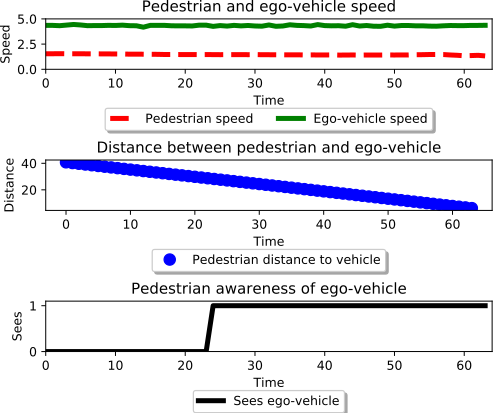}}
\caption{Each column represents an example of different types of interactions between pedestrian and ego-vehicle. First and second columns exemplify scenarios where pedestrian and ego-vehicle affect each other's decisions. The third scenario both agents have not strongly affected each other's actions. In each column the first image is the scenario with the pedestrian highlighted in a red bounding box, the three graphs behind each scenario represent pedestrian and ego-vehicle velocities, the distance between pedestrian and ego-vehicle, and the time-steps in which pedestrian looks at the ego-vehicle direction (1 = pedestrian sees vehicle, 0 = pedestrian does not see vehicle).}
\label{fig:differentcases}
\end{figure*}

In order to understand better such interactions between pedestrians and cars we used the data provided in \cite{Rehder2018} that contains realistic inner-city scenarios with data collected by a moving vehicle. The pedestrians were manually annotated in all sequences, and their trajectories were optimized from stereo imaging using a constant velocity model. The dataset contains \gls{GT} labels for pedestrians' positions, head orientation and occlusion information. A human annotator watched the videos from the dataset and labeled all the sequences where the pedestrian intended to walk through the ego-vehicle lane. 
From those sequences we analyzed interactions based on the pedestrian and car dynamics, and  pedestrian awareness (and possible eye contact) estimated by head orientation. As the dataset did not comprise images from the driver (inside car images) we were restricted to analyze the visual interactions by the pedestrian side.

After labelling the sequences in which the pedestrians intend to cross the ego-vehicle lane, we split the sequences into validation, test and train. For train and validation sequences we also used sequences from the dataset where the pedestrians were not intending to cross the street. 
We subdivided the sequences using a fixed length and a frame rate of 5Hz, i\,.e.  $0.2$ second interval between each input, (the original frame rate of the dataset in \cite{Rehder2018} was $10 Hz$). In order to augment the data we used a sliding window of $1$ time step (varying $0.1$ seconds). We used a total of trajectories of 8941 for train, 3652 for validation and 834 for test. We used a frame of reference with the origin fixed at the time $t$ of pedestrian being predicted.

In Fig. \ref{fig:differentcases} we present three scenarios found in the data. The first one (Fig. \ref{fig:seekapprovalfig}) the pedestrian decrease his own speed and seek for eye contact with the driver, the driver then decided to decrease the ego-vehicle speed, thus encouraging the pedestrian to cross. The graphs in Fig \ref{fig:seekapprovalfigb} display such behavior through the collected data, i\,.e. in first graph between $T=20$ and $T=50$ the driver decreases the car's speed and the pedestrian increases his own speed, therefore crossing the street. The second scenario (Fig. \ref{fig:pedstopfig}) the  opposite happens, the car did not stop so the pedestrian is forced to stop at the curb. This two scenarios display cases where the ego-vehicle and pedestrian decisions/actions are affected by each other. On the first scenario the pedestrian is encouraged to cross and on the second the pedestrian is forced to stop. This supports the importance of studying such interactions and how they affect decision. The third figure (Fig. \ref{fig:highdistanceped}) displays a scenario where the ego-vehicle and pedestrian do not affect each other's decisions. In such scenario the pedestrian is already crossing the street and he is at a high distance from the ego-vehicle.

On the table \ref{tab:results} we present the RMSE loss comparison among the different experiments:

\begin{itemize}
    \item \textbf{Baseline}: Pedestrian past trajectory
    \item \textbf{Method 1}: Pedestrian and ego-vehicle past trajectories
    \item \textbf{Method 2}: Pedestrian and ego-vehicle past trajectories and pedestrian head orientation
\end{itemize}

\begin{figure*}
\centering
\subfigure[Model using only pedestrian past positions (blue) RMSE = 0.58. Model using pedestrian and ego-vehicle past positions (red) RMSE=0.38]{\label{fig:ped1}\includegraphics[width=8cm]{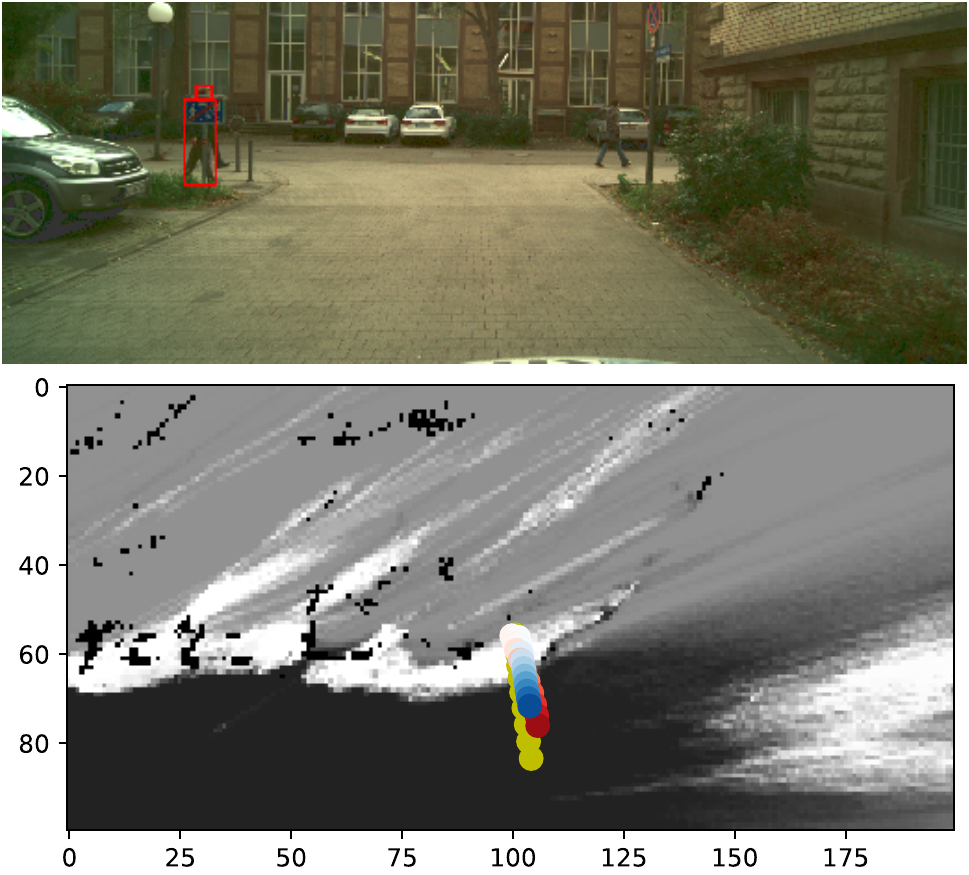}}
\qquad
\subfigure[Model using only pedestrian past positions (blue) RMSE = 0.81. Model using pedestrian and ego-vehicle past positions, and pedestrian head orientation (red) RMSE=0.50.]{\label{fig:ped2}\includegraphics[width=8cm]{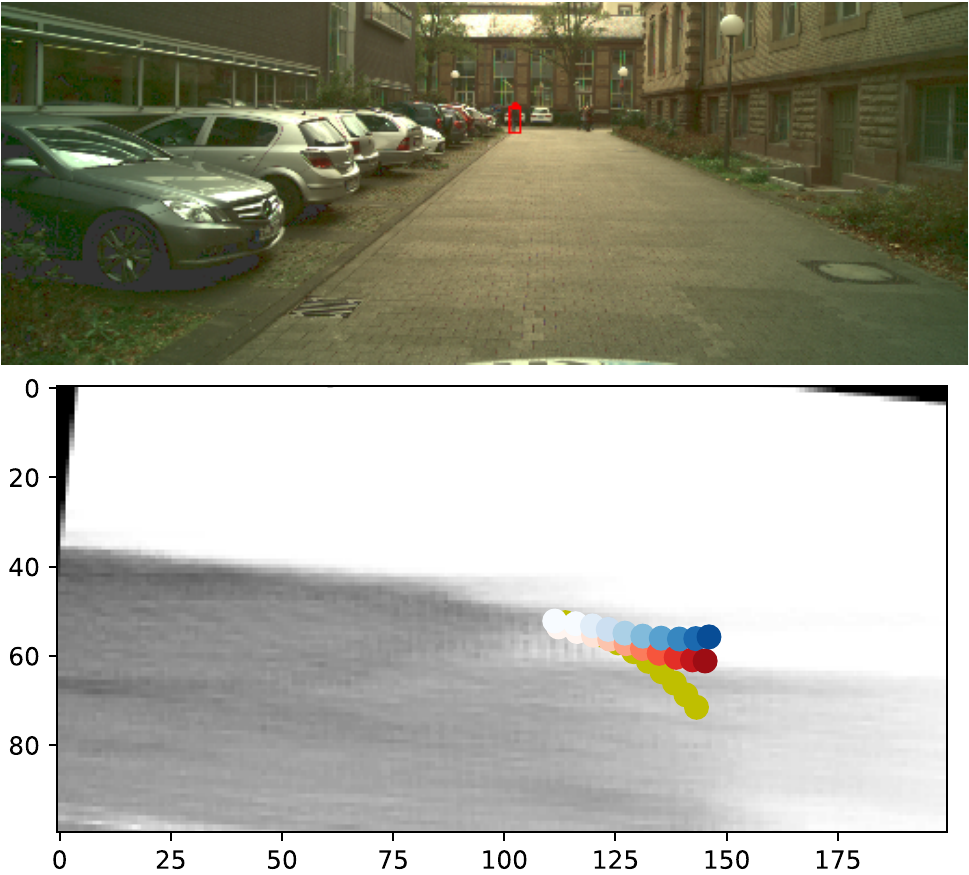}}
\caption{Comparison between the model that uses only pedestrian positions (blue circles) and the model that uses also ego-vehicle positions and pedestrian head orientation (red circles), yellow circles represents GT pedestrian position. \gls{BEV} map white color represents sidewalk and black represent the street. It is possible to see that the output given by the model that takes into consideration head and ego-vehicle information got closer to the \gls{GT} positions. This visual understanding is also reflected in the RMSE values (fig. \ref{fig:rmsefig}).}
\label{fig:ped_results}
\end{figure*}

Figure \ref{fig:ped_results} shows qualitative results for two sequences in which pedestrian intends to cross a street. At the top row, the image has the pedestrian highlighted in a red bounding box, and at the bottom image the \gls{BEV} map presents the qualitative results together with the \gls{GT}, where color white represents sidewalk, black street, yellow circles represent pedestrian \gls{GT} positions, red circles represent pedestrian predicted positions by the proposed model, and blue circles represent the baseline model using only pedestrian past positions. From both images and RMSE errors we can see that the proposed method got closer to the \gls{GT} positions, however, in scenarios where this extra information does not affect pedestrian decision, e\,.g. after a decision has been made and the pedestrian start moving, both models tend to perform similarly,  without a considerable improvement from the model that uses the interaction cues.

Dealing only with scenarios where the pedestrian intended to walk though ego-lane decreased considerably the amount of data. We believe with more data the network would learn to rely more on such patterns of interactions. We could also notice that some pedestrians that were unaware of the ego-vehicle were exhibiting a follow-me behavior, i\,.e. relying in the perception of a "leader".

\begin{table}
\caption{RMSE loss comparison among different experiments}
\begin{center}
\begin{tabular}{ c|c|c}    \toprule
& \emph{input} & \emph{RMSE} (meters)\\\hline
Baseline &ped. pos. only    & 0.53 \\
Method 1 &ped. and veh. pos  & 0.443\\
Method 2 &ped. and veh. pos, and head orientation & 0.427\\\bottomrule
 \hline
\end{tabular}
\end{center}
\label{tab:results}
\end{table}

\begin{figure}
\centering
\includegraphics[width=\columnwidth]{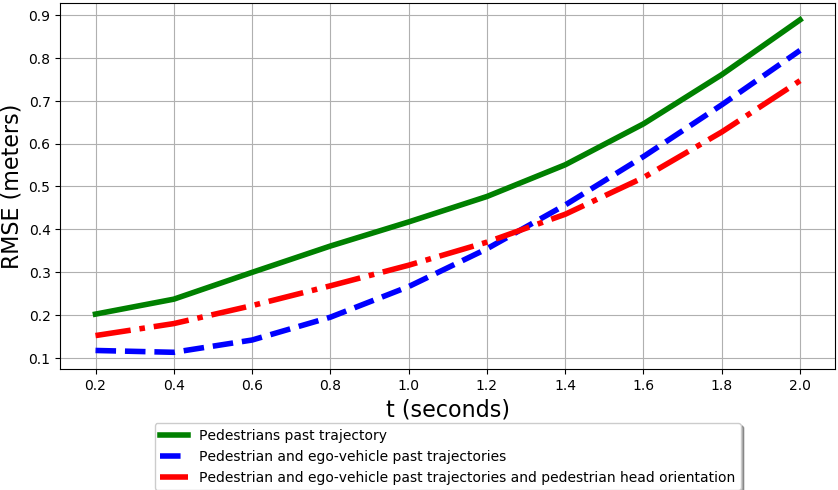}
\caption{RMSE comparison between baseline and proposed approach.}
\label{fig:rmsefig}
\end{figure}

\section{Discussion and Future Work}
This paper provided an analysis of pedestrian and driver interactions through dynamics and awareness using naturalistic data (i\,.e. real inner-city data where the pedestrians are not actors performing pre-defined actions).  It was possible to observe that pedestrians decisions are affected by other agents. As our data only covered the ego-vehicle and pedestrians, we were constrained to study only such interaction. A broader study regarding the inclusion of other traffic agents is still necessary, as pedestrians decisions might also be affected by other pedestrians and cars in the scene.

Such interaction through means of speed and eye contact usually affects pedestrian trajectory when he/she aims at crossing ego-vehicle lane. Given that such scenarios are only a portion of the available datasets more work has to be done in collecting more diverse scenarios that comprise different pedestrians and actions.

Future work will focus on incorporating information from other agents, environment, and different pedestrian types. We also would like to model pedestrian future trajectories in a spatial compliance with scene as the pedestrian future trajectory is bounded by static objects (buildings) and different types of surfaces, i\,.e. pedestrians may give preference to sidewalk over street.

Pedestrians and drivers are constantly interacting, algorithms that predict pedestrian future positions or actions should take into consideration ego-vehicle motion data as pedestrians take decisions using such information. Current available datasets usually have data collected with a car driven by a human, such interactions going on between pedestrians and the drivers should also be stored and used to improve robustness of current methods for pedestrian path prediction, as a head yielding preference might be a strong cue that is not captured by outside sensors.

Likewise, self-driving cars should be able to incorporate behaviors that are used by cars nowadays on the streets. Beyond gestures and eye-contact, patterns of motion of both agents are key aspects that interferes in other agents behaviors on streets.

The  human/machine interface between autonomous cars and pedestrians is also something that will interfere with pedestrians behavior in the future. Such interfaces should be adaptable to the different types of pedestrians moving around on cities. The way the car will communicate is also very dependent on whom it is interacting with. Most of the conceptual projects proposed by automotive companies and some academic researches focus on the autonomous car displaying messages. However, this communication should also be able to adapt to the needs of the pedestrian.

\section*{Acknowledgement}
We would like to thank Eike Rehder, Florian Wirth and Jannik Quehl from MRT-KIT group in Germany for the data used in this work. We also would like to thank Fulbright and CAPES for the financial support in this research.
This study was financed in part by the Coordenaç\~{a}o de Aperfeiçoamento de Pessoal de N\'{i}vel Superior - Brasil (CAPES) - Finance Code 001.

\bibliographystyle{IEEEtran}
\bibliography{refs}

\begin{thebibliography}{10}
\providecommand{\url}[1]{#1}
\csname url@rmstyle\endcsname
\providecommand{\newblock}{\relax}
\providecommand{\bibinfo}[2]{#2}
\providecommand\BIBentrySTDinterwordspacing{\spaceskip=0pt\relax}
\providecommand\BIBentryALTinterwordstretchfactor{4}
\providecommand\BIBentryALTinterwordspacing{\spaceskip=\fontdimen2\font plus
\BIBentryALTinterwordstretchfactor\fontdimen3\font minus
  \fontdimen4\font\relax}
\providecommand\BIBforeignlanguage[2]{{%
\expandafter\ifx\csname l@#1\endcsname\relax
\typeout{** WARNING: IEEEtran.bst: No hyphenation pattern has been}%
\typeout{** loaded for the language `#1'. Using the pattern for}%
\typeout{** the default language instead.}%
\else
\language=\csname l@#1\endcsname
\fi
#2}}

\bibitem{who2}
W.~H.~O. WHO, ``Global status report on road safety,'' 2015, p.~8.

\bibitem{Rasouli2018itsc}
A.~Rasouli, I.~Kotseruba, and J.~K. Tsotsos, ``Towards social autonomous
  vehicles: Understanding pedestrian-driver interactions,'' in \emph{2018 21st
  International Conference on Intelligent Transportation Systems (ITSC)}, Nov
  2018, pp. 729--734.

\bibitem{Rothenbucher2016}
D.~Rothenbücher, J.~Li, D.~Sirkin, B.~Mok, and W.~Ju, ``Ghost driver: A field
  study investigating the interaction between pedestrians and driverless
  vehicles,'' in \emph{2016 25th IEEE International Symposium on Robot and
  Human Interactive Communication (RO-MAN)}, Aug 2016, pp. 795--802.

\bibitem{Gupta2018}
S.~Gupta, M.~Vasardani, and S.~Winter, ``Negotiation between vehicles and
  pedestrians for the right of way at intersections,'' \emph{IEEE Transactions
  on Intelligent Transportation Systems}, pp. 1--12, 2018.

\bibitem{Chang2017EyesOA}
C.-M. Chang, K.~Toda, D.~Sakamoto, and T.~Igarashi, ``Eyes on a car: an
  interface design for communication between an autonomous car and a
  pedestrian,'' in \emph{AutomotiveUI}, 2017.

\bibitem{dey2017}
D.~Dey and J.~Terken, ``Pedestrian interaction with vehicles: roles of explicit
  and implicit communication,'' in \emph{Proceedings of the 9th International
  Conference on Automotive User Interfaces and Interactive Vehicular
  Applications}.\hskip 1em plus 0.5em minus 0.4em\relax ACM, 2017, pp.
  109--113.

\bibitem{schneemann2016}
F.~Schneemann and I.~Gohl, ``Analyzing driver-pedestrian interaction at
  crosswalks: A contribution to autonomous driving in urban environments,'' in
  \emph{Intelligent Vehicles Symposium (IV), 2016 IEEE}.\hskip 1em plus 0.5em
  minus 0.4em\relax IEEE, 2016, pp. 38--43.

\bibitem{Wang2016}
C.~Wang, A.~Liu, P.~Wu, and P.~Lu, ``A study in human-machine interaction
  through agent simulation: An application in pedestrian crossing,'' in
  \emph{2016 International Automatic Control Conference (CACS)}, Nov 2016, pp.
  167--172.

\bibitem{Gupta2016}
\BIBentryALTinterwordspacing
S.~Gupta, M.~Vasardani, and S.~Winter, ``Conventionalized gestures for the
  interaction of people in traffic with autonomous vehicles,'' in
  \emph{Proceedings of the 9th ACM SIGSPATIAL International Workshop on
  Computational Transportation Science}, ser. IWCTS '16.\hskip 1em plus 0.5em
  minus 0.4em\relax New York, NY, USA: ACM, 2016, pp. 55--60. [Online].
  Available: \url{http://doi.acm.org/10.1145/3003965.3003967}
\BIBentrySTDinterwordspacing

\bibitem{Rangesh2016}
A.~Rangesh, E.~Ohn-Bar, K.~Yuen, and M.~M. Trivedi, ``Pedestrians and their
  phones - detecting phone-based activities of pedestrians for autonomous
  vehicles,'' in \emph{2016 IEEE 19th International Conference on Intelligent
  Transportation Systems (ITSC)}, Nov 2016, pp. 1882--1887.

\bibitem{Rasouli2018}
``Autonomous vehicles that interact with pedestrians: A survey of theory and
  practice,'' \url{https://arxiv.org/abs/1805.11773}, accessed: 2019-01-23.

\bibitem{Pillai2017}
A.~Pillai, ``{KTH Master Thesis: Virtual Reality based Study to Analyse
  Pedestrian attitude towards Autonomous Vehicles},'' Master's thesis, KTH,
  Stockholm, 2017.

\bibitem{Ridel2018}
D.~Ridel, E.~Rehder, M.~Lauer, C.~Stiller, and D.~Wolf, ``A literature review
  on the prediction of pedestrian behavior in urban scenarios,'' in \emph{2018
  21st International Conference on Intelligent Transportation Systems (ITSC)},
  Nov 2018, pp. 3105--3112.

\bibitem{Alahi2016}
A.~Alahi, K.~Goel, V.~Ramanathan, A.~Robicquet, L.~Fei-Fei, and S.~Savarese,
  ``Social {LSTM}: Human trajectory prediction in crowded spaces,'' in
  \emph{2016 IEEE Conference on Computer Vision and Pattern Recognition
  (CVPR)}, June 2016, pp. 961--971.

\bibitem{kitani2012eccv}
K.~M. Kitani, B.~D. Ziebart, J.~A. Bagnell, and M.~Hebert, ``Activity
  forecasting,'' in \emph{Computer Vision -- ECCV 2012: 12th European
  Conference on Computer Vision, Florence, Italy, October 7-13, 2012,
  Proceedings, Part IV}, A.~Fitzgibbon, S.~Lazebnik, P.~Perona, Y.~Sato, and
  C.~Schmid, Eds.\hskip 1em plus 0.5em minus 0.4em\relax Berlin, Heidelberg:
  Springer Berlin Heidelberg, 2012, pp. 201--214.

\bibitem{socgan}
A.~Gupta, J.~Johnson, L.~Fei-Fei, S.~Savarese, and A.~Alahi, ``Social gan:
  Socially acceptable trajectories with generative adversarial networks,'' in
  \emph{IEEE Conference on Computer Vision and Pattern Recognition (CVPR)}, no.
  CONF, 2018.

\bibitem{desire}
N.~Lee, W.~Choi, P.~Vernaza, C.~B. Choy, P.~H. Torr, and M.~Chandraker,
  ``Desire: Distant future prediction in dynamic scenes with interacting
  agents,'' in \emph{Proceedings of the IEEE Conference on Computer Vision and
  Pattern Recognition}, 2017, pp. 336--345.

\bibitem{carnet}
A.~Sadeghian, F.~Legros, M.~Voisin, R.~Vesel, A.~Alahi, and S.~Savarese,
  ``Car-net: Clairvoyant attentive recurrent network,'' in \emph{Proceedings of
  the European Conference on Computer Vision (ECCV)}, 2018, pp. 151--167.

\bibitem{sophie}
A.~Sadeghian, V.~Kosaraju, A.~Sadeghian, N.~Hirose, and S.~Savarese, ``Sophie:
  An attentive gan for predicting paths compliant to social and physical
  constraints,'' \emph{arXiv preprint arXiv:1806.01482}, 2018.

\bibitem{vemula2018}
A.~Vemula, K.~Muelling, and J.~Oh, ``Social attention: Modeling attention in
  human crowds,'' in \emph{2018 IEEE International Conference on Robotics and
  Automation (ICRA)}.\hskip 1em plus 0.5em minus 0.4em\relax IEEE, 2018, pp.
  1--7.

\bibitem{varshneya2017}
D.~Varshneya and G.~Srinivasaraghavan, ``Human trajectory prediction using
  spatially aware deep attention models,'' \emph{arXiv preprint
  arXiv:1705.09436}, 2017.

\bibitem{cheng2018}
H.~Cheng and M.~Sester, ``Modeling mixed traffic in shared space using lstm
  with probability density mapping,'' in \emph{2018 21st International
  Conference on Intelligent Transportation Systems (ITSC)}.\hskip 1em plus
  0.5em minus 0.4em\relax IEEE, 2018, pp. 3898--3904.

\bibitem{fernando2018}
T.~Fernando, S.~Denman, S.~Sridharan, and C.~Fookes, ``Soft+ hardwired
  attention: An lstm framework for human trajectory prediction and abnormal
  event detection,'' \emph{Neural networks}, vol. 108, pp. 466--478, 2018.

\bibitem{ballan2016}
L.~Ballan, F.~Castaldo, A.~Alahi, F.~Palmieri, and S.~Savarese, ``Knowledge
  transfer for scene-specific motion prediction,'' in \emph{European Conference
  on Computer Vision}.\hskip 1em plus 0.5em minus 0.4em\relax Springer, 2016,
  pp. 697--713.

\bibitem{brendan2018}
N.~Nikhil and B.~T. Morris, ``Convolutional neural network for trajectory
  prediction,'' in \emph{European Conference on Computer Vision}.\hskip 1em
  plus 0.5em minus 0.4em\relax Springer, 2018, pp. 186--196.

\bibitem{hasan2018mx}
I.~Hasan, F.~Setti, T.~Tsesmelis, A.~Del~Bue, F.~Galasso, and M.~Cristani,
  ``Mx-lstm: mixing tracklets and vislets to jointly forecast trajectories and
  head poses,'' \emph{arXiv preprint arXiv:1805.00652}, 2018.

\bibitem{Schulz2015}
A.~T. Schulz and R.~Stiefelhagen, ``A controlled interactive multiple model
  filter for combined pedestrian intention recognition and path prediction,''
  in \emph{2015 IEEE 18th International Conference on Intelligent
  Transportation Systems}, 2015, pp. 173--178.

\bibitem{Schulz2015b}
Schulz and R.~Stiefelhagen, ``Pedestrian intention recognition using
  latent-dynamic conditional random fields,'' in \emph{2015 IEEE Intelligent
  Vehicles Symposium (IV)}, 2015, pp. 622--627.

\bibitem{Kooij2014}
J.~F.~P. Kooij, N.~Schneider, F.~Flohr, and D.~M. Gavrila, ``Context-based
  pedestrian path prediction,'' in \emph{Computer Vision -- ECCV 2014: 13th
  European Conference, Zurich, Switzerland, September 6-12, 2014, Proceedings,
  Part VI}, D.~Fleet, T.~Pajdla, B.~Schiele, and T.~Tuytelaars, Eds.\hskip 1em
  plus 0.5em minus 0.4em\relax Cham: Springer International Publishing, 2014,
  pp. 618--633.

\bibitem{Hashimoto2015b}
Y.~Hashimoto, Y.~Gu, L.~T. Hsu, and S.~Kamijo, ``Probability estimation for
  pedestrian crossing intention at signalized crosswalks,'' in \emph{2015 IEEE
  International Conference on Vehicular Electronics and Safety (ICVES)}, 2015,
  pp. 114--119.

\bibitem{Hashimoto2015}
Y.~Hashimoto, G.~Yanlei, L.~T. Hsu, and K.~Shunsuke, ``A probabilistic model
  for the estimation of pedestrian crossing behavior at signalized
  intersections,'' in \emph{2015 IEEE 18th International Conference on
  Intelligent Transportation Systems}, 2015, pp. 1520--1526.

\bibitem{Keller2014}
C.~G. Keller and D.~M. Gavrila, ``Will the pedestrian cross? a study on
  pedestrian path prediction,'' \emph{IEEE Transactions on Intelligent
  Transportation Systems}, vol.~15, no.~2, pp. 494--506, 2014.

\bibitem{semconreport}
Semcon, ``Self-driving car smiles at pedestrians,'' in \emph{Attitudes to
  self-driving cars}, 2016.

\bibitem{Deo2018}
N.~Deo and M.~M. Trivedi, ``Convolutional social pooling for vehicle trajectory
  prediction,'' in \emph{2018 IEEE/CVF Conference on Computer Vision and
  Pattern Recognition Workshops (CVPRW)}, June 2018, pp. 1549--15\,498.

\bibitem{Paszke2017pytorch}
A.~Paszke, S.~Gross, S.~Chintala, G.~Chanan, E.~Yang, Z.~DeVito, Z.~Lin,
  A.~Desmaison, L.~Antiga, and A.~Lerer, ``Automatic differentiation in
  pytorch,'' 2017.

\bibitem{manualpytorch}
``Reproducibility,''
  \url{https://pytorch.org/docs/stable/notes/randomness.html}, accessed:
  2019-01-25.

\bibitem{kingma2014adam}
D.~P. Kingma and J.~Ba, ``Adam: A method for stochastic optimization,''
  \emph{arXiv preprint arXiv:1412.6980}, 2014.

\bibitem{Rehder2018}
E.~Rehder, F.~Wirth, M.~Lauer, and C.~Stiller, ``Pedestrian prediction by
  planning using deep neural networks,'' in \emph{2018 IEEE International
  Conference on Robotics and Automation (ICRA)}, May 2018, pp. 1--5.

\end{thebibliography}

\end{document}